\documentclass{article}

\usepackage[dblblindworkshop, final]{neurips_2025}
\usepackage{float}
\workshoptitle{CogInterp Interpreting Cognition
in Deep Learning Models}

\usepackage[utf8]{inputenc} % allow utf-8 input
\usepackage[T1]{fontenc}    % use 8-bit T1 fonts
\usepackage{hyperref}       % hyperlinks
\usepackage{url}            % simple URL typesetting
\usepackage{booktabs}       % professional-quality tables
\usepackage{subcaption}    % for subfigure environment
\usepackage{amsfonts}       % blackboard math symbols
\usepackage{nicefrac}       % compact symbols for 1/2, etc.
\usepackage{microtype}      % microtypography
\usepackage{xcolor}         % colors

\title{Modulation of temporal decision-making in a deep reinforcement learning agent under the dual-task paradigm }

\author{%
  Amrapali Pednekar \\
    Department of Information Technology \\
    IDLab, Ghent University - imec \\
  \texttt{Amrapali.Pednekar@UGent.be} \\
  \And
  \'Alvaro Garrido-P\'erez \\
    Department of Information Technology \\
    IDLab, Ghent University - imec \\ 
  \texttt{Alvaro.GarridoPerez@UGent.be} \\
  \AND
  Yara Khaluf \\
  Department of Social Sciences\\
  Wageningen University and Research \\
  \texttt{yara.khaluf@wur.nl} \\
  \And
   Pieter Simoens \\
  Department of Information Technology \\
    IDLab, Ghent University - imec \\ 
  \texttt{Pieter.Simoens@UGent.be} \\
}

\usepackage{graphicx} % Required for inserting images

\begin{document}

\maketitle

\begin{abstract}
  This study explores the interference in temporal processing within a dual-task paradigm from an artificial intelligence (AI) perspective. In this context, the dual-task setup is implemented as a simplified version of the Overcooked environment with two variations, single task (T) and dual task (T+N). Both variations involve an embedded time production task, but the dual task (T+N) additionally involves a concurrent number comparison task. Two deep reinforcement learning (DRL) agents were separately trained for each of these tasks. These agents exhibited emergent behavior consistent with human timing research. Specifically, the dual task (T+N) agent exhibited significant overproduction of time relative to its single task (T) counterpart. This result was consistent across four target durations. Preliminary analysis of neural dynamics in the agents' LSTM layers did not reveal any clear evidence of a dedicated or intrinsic timer. Hence, further investigation is needed to better understand the underlying time-keeping mechanisms of the agents and to provide insights into the observed behavioral patterns. This study is a small step towards exploring parallels between emergent DRL behavior and behavior observed in biological systems in order to facilitate a better understanding of both. 
\end{abstract}

\section{Introduction}

How do humans track time? This question has captivated researchers in psychology, neuroscience, and cognitive science for many years \cite{grondin2010timing,block2014time, buhusi2005makes, merchant2013neural, basgol2021time}. 
Despite extensive research, there is a fragmented and sometimes conflicting understanding of temporal processing and its underlying mechanisms. However, parallel research across different fields has led to convergence on certain aspects of timing \cite{matthews2014time,eagleman2008human}. This reflects both the inherent complexity of the problem and the importance of incorporating complementary tools and perspectives to further deepen our understanding of human timing. 

The current study attempts to touch on this problem using an artificial intelligence (AI) perspective. To do so, we consider a well-established and robust finding from timing research. Namely, the modulation of time perception in a dual-task paradigm. Numerous behavioral studies have shown that timing performance is affected by the presence of a concurrent cognitive task \cite{brown2013timing, brown1997attentional, brown2006timing, fortin1987time,nobre2010attention}. Such findings have implicated shared resources between temporal and cognitive processes. Furthermore, functional brain imaging studies have shown that temporal processing is distributed across multiple brain regions, and that neurons involved in timing also participate in other cognitive functions \cite{merchant2013neural,buhusi2005makes}. Thus, time is affected by concurrent cognitive tasks possibly because temporal processing is encoded in the activity of neurons that are also involved in other cognitive processes. 

% result 1
We show that deep reinforcement learning (DRL) agents exhibit behavior that suggests temporal interference in a dual-task paradigm. Specifically, two DRL agents were trained separately on a single task and a dual task. The single task variation involved an embedded time production task (referred as single task (T)), while the dual task variation involved the time production task and a concurrent number comparison task (referred as dual task (T+N)).  All other task parameters and agent characteristics were kept the same across the two tasks. To substantiate the findings, training was performed separately for four target durations (7, 8, 9, and 10 time steps), resulting in four independently trained agents per task type. Analysis of the distribution of durations produced by the different agents revealed that dual task (T+N) DRL agents significantly overproduced across all durations as compared to their single task (T) counterparts.  Thus, the emergent behavior observed in DRL agents under a dual-task paradigm parallels findings from behavioral studies on time production with interference \cite{fortin1987time,brown1997attentional,brown2006timing}. 

% result 2
A preliminary analysis of neural dynamics in both types of agents did not show strong evidence for a dedicated  or an intrinsic timer \cite{ivry2008dedicated}. For some target durations, oscillations with frequencies equal to the target duration were observed in the latent space of neural activities. However, further analysis is needed to confirm that these oscillations correspond to a time-keeping mechanism. Thus, no conclusive neural evidence was found to explain the observed behavioral differences.

\section{Methodology}

\begin{figure}[htbp]
    \centering
    \begin{subfigure}[t]{0.4\textwidth}
        \centering
        \vspace{-6cm}
        \includegraphics[width=\textwidth]{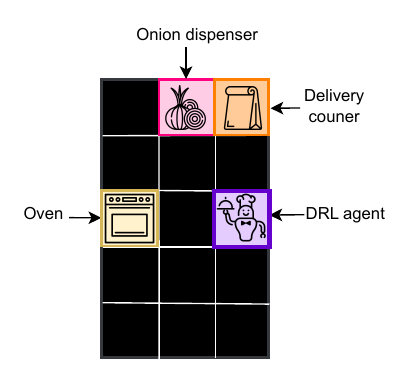}
        \vspace{0.6cm}
        \subcaption{Single task (T)}
        \label{fig:grid_singletask}
    \end{subfigure}
    \hfill
    \begin{subfigure}[t]{0.4\textwidth}
        \centering
        \includegraphics[width=\textwidth]{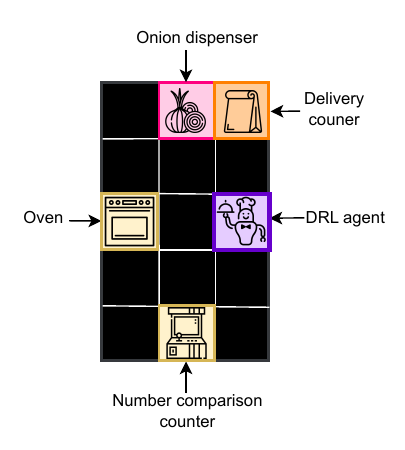}
         % \vspace{-0.1cm}
        \subcaption{Dual task (T+N)}
        \label{fig:grid_dualtask}
    \end{subfigure}
    \caption{Grid worlds representing a simplified version of the OverCooked environment \cite{carroll2019utility}, used for the single task (T) and dual task (T+N) experiments (icons were sourced from Flaticon.com).}
    \label{fig:grid_overall}
\end{figure}

The Reinforcement learning (RL) environment was a simplified version of the popular OverCooked environment \cite{carroll2019utility}. The modified setup (shown in Figure \ref{fig:grid_singletask}) consisted of a single-agent scenario with a 5x3 grid world featuring three counters (an onion dispenser, an oven and a delivery counter) and two objects (onion and soup). The agent could perform six actions, five for navigation (`wait', `up', `down', `left' and `right') and one `interact' action to engage with an adjacent counter or object. 
% There were no obstructions for the agent's movements, except that it could not move through the counters.

% make better - soup can be overcooked 
The agent’s goal was to deliver soup to the delivery counter. To do this, it had to pick up an onion from the dispenser, place it in the oven, and wait for the soup to cook. The oven started an internal (invisible) timer upon receiving the onion, tied to a target duration. The soup became ready only after this duration and thus interacting with the oven before that had no effect. Once ready, the agent could take the soup at any point (i.e., at or after the target duration) and deliver it to complete a trial. Each soup delivery yielded a `+1' reward, while all other actions yielded zero reward, even if the soup stayed in the oven beyond the target time. The agent had to deliver as many soups as possible per episode (consisting of 100 time steps), making accurate timing essential for optimal performance.

% For example, the agent `interacts' with onion dispenser by being in one of the adjacent blocks and taking the `interact' actions to get an onion. Similarly, it puts the onion in the oven or takes the soup out of the oven only after interacting with the oven in addition to standing in one of the adjacent blocks. Same goes for delivering soup at the delivery counter. 

In the dual-task variation, a number comparison counter was added to the above setup (shown in Figure \ref{fig:grid_dualtask}). After placing the onion in the oven, this counter activated for 4 time steps (irrespective of the target duration), displaying numbers between 1 and 10. The agent had to respond with the `interact' action if the number was less than 5, or the `wait' action if it was 5 or more. It received an immediate `+1' reward for correctly performing the number comparison task and a zero reward otherwise. After four time steps, the number comparison ended, and the task proceeded as in the single task (T) setup. It is important to note that, the agent always had enough time in all target intervals to retrieve the soup at least one time step before the target duration.

% Thus in addition to learnign to deliver the soups the agent had to keep a track of the oven timer. Reaching the oven at the target interval was optimal because it would mean more soup delivery. In the dual-task variation, in addition to doing the soup deliver the agent also had to discover the number game block and correctly perform the number task to get additional reward. Thus this setup is more than just a pure interval timing task. The agent has to navigate and learn to do that task while having to keep a track of time for a certain aspect of the task. Through this setup we aim to study how an additional task affects the timing performance of the agent. And whether there is a difference in how the agent keeps a track of time in both cases. 

% add architecture details to apendix 
The DRL agents were implemented using the recurrent policy variant of the Proximal Policy Optimization (PPO) algorithm from the Stable-Baselines3 (SB3) library \cite{stable-baselines3} (see Appendix). An entropy coefficient of 0.05 encouraged exploration, causing the agent to move around the grid while the soup was cooking. This proved especially helpful in successfully training the dual-task variation. For both tasks, the corresponding agents were trained for 100,000 time steps, which was sufficient to learn the tasks and achieve comparable performance.

% Since th grid is small, the CNN most consisted of 1*1 convolutions to process spatial data. . 

% define trial, episode , time steps, oven timer 

\begin{figure}[h!]
  \centering
  \includegraphics[width=0.85\textwidth]{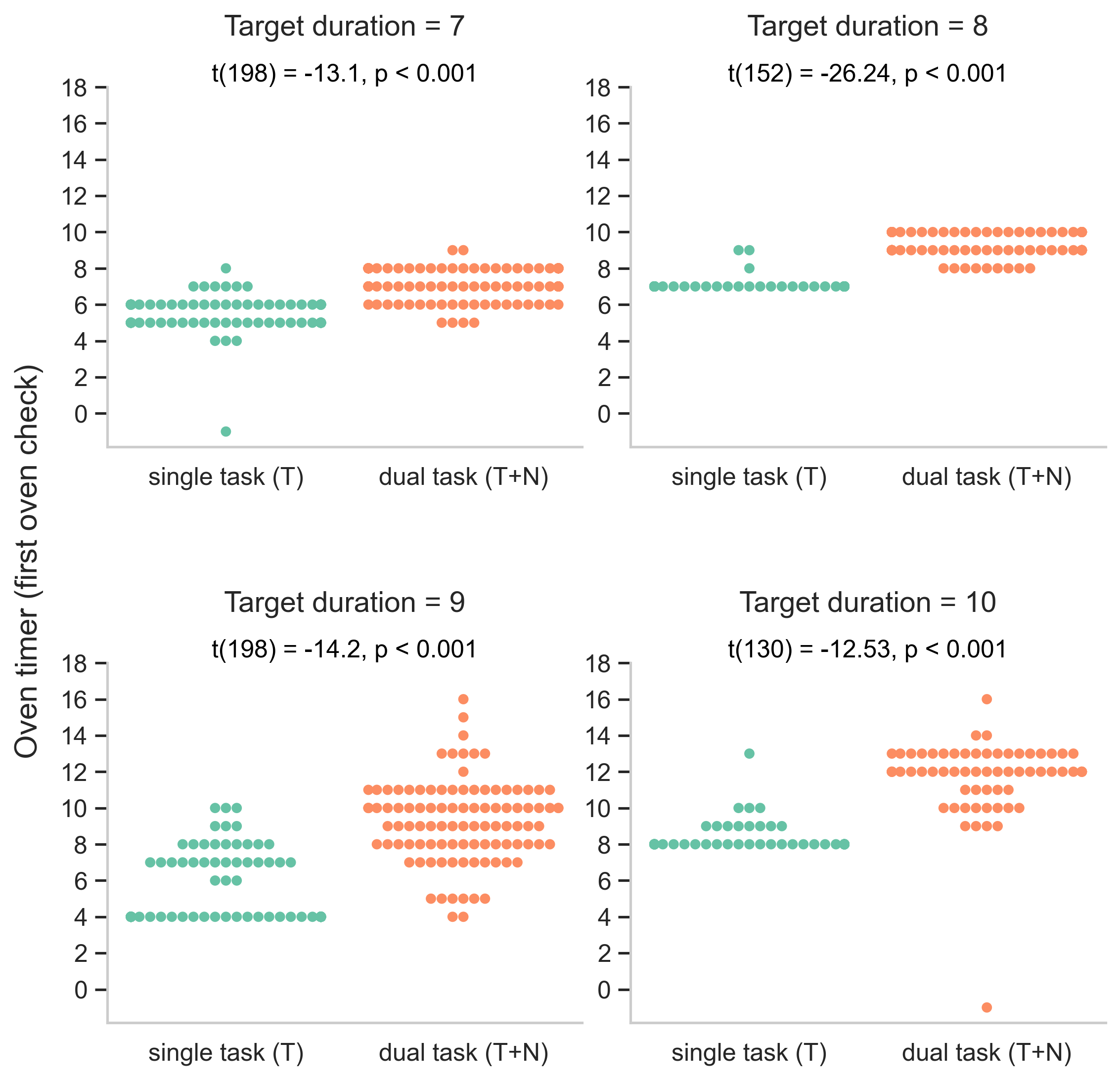}
  \caption{Distribution of oven timers corresponding to the `first oven check' in 25 episodes (each of 100 time steps) across the two task types, shown for different target durations. The dual task (T+N) agent tends to significantly overestimate (p < 0.001) as compared to its single task (T) counterpart. The corresponding independent t-test statistics are shown in the plots. }
  \label{fig:first_oven_check}
\end{figure}

\section{Behavioral analysis}

% first oven check definition 
For the behavioral analysis, we define the `first oven check' as the oven timer value at which the agent initiates the `Interact' action with an onion carrying oven. If the oven timer is more than or equal to the target duration, agent can take the soup out and the corresponding oven time is recorded as the `first oven check' for that trial. In contrast, if the oven timer is less than the target duration, the agent cannot take the soup out. In this case, if the agent consecutively continues to `Interact' with the oven until it can take the soup out (i.e., until the target duration), the oven timer corresponding to the first `Interact' action is recorded as the the `first oven check'. However, if at least one of the consecutive actions is not `Interact' the oven timer is not recorded. The intuition is that the agent checks the oven because it considers the target duration to be reached and continues to check it until it can get the soup out.  Thus, by comparing the `first oven checks' for both tasks, we can assess whether the concurrent number comparison game induced a change in timing behavior.

% results 
Figure \ref{fig:first_oven_check} shows the distribution of `first oven checks' across 25 episodes (each consisting of 100 time steps) for the different target durations. It can be seen that the average `first oven check' is significantly higher (p<0.001) in the dual task (T + N) as compared to the single task (T). The independent t-test statistics and degrees of freedom are shown in Figure \ref{fig:first_oven_check}. Thus, on average, the dual task (T+N) agents exhibit a significant overproduction of time relative to their single task (T) counterparts. 

\begin{figure}[h!]
  \centering
  \includegraphics[width=\textwidth]{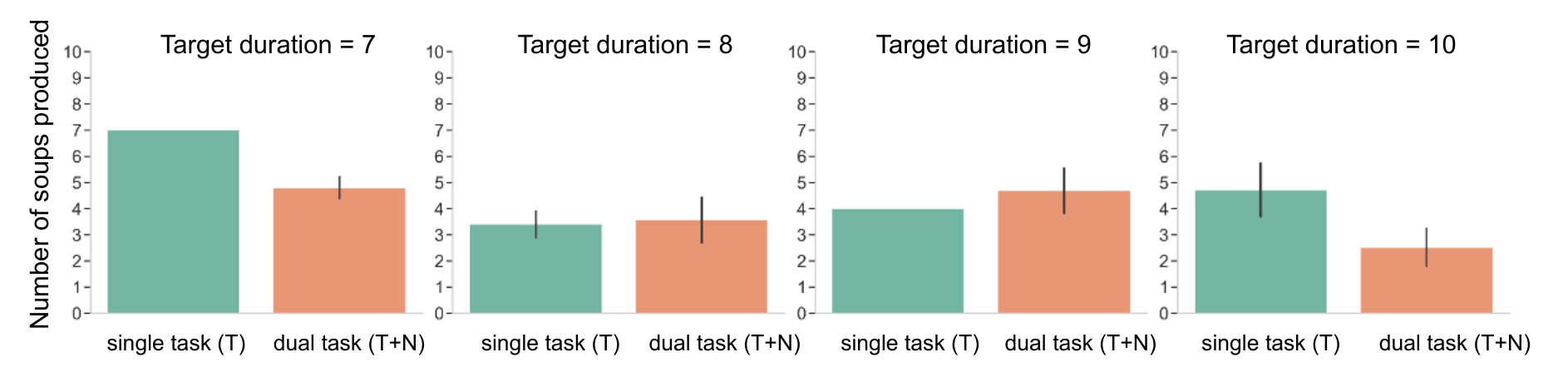}
  \caption{Average number of soups produced across the 25 episodes (each of 100 time steps) for the two task types, shown for different target durations. }
  \label{fig:no_soups_produced}
\end{figure}

It is important to note that in the dual tasks, the agents had enough time after the number game stopped to go to the oven and `Interact' with it at least one time step before the target duration. Moreover, despite being trained for the same number of training steps, the dual task (T+N) agents achieve performance comparable to their single-task (T) counterparts. Figure \ref{fig:no_soups_produced} shows the average number of soups produced across the 25 episodes in the two task types, for different target durations. Since an agent can only produce a new soup after delivering the previous one,  Figure \ref{fig:no_soups_produced} reflects agent performance. Even in the least favorable case (target duration 10), the dual task (T+N) agent produces \textasciitilde 53\% of the soups produced by its single task (T) counterpart.

% conclusion - lr about timing bias in artifiical agents 
Thus, behaviorally, the DRL agents seem to replicate the well-established findings in human timing, namely, overproduction of time in the presence of a concurrent cognitive task \cite{fortin1987time,brown1997attentional,brown2006timing}. It is important to note that this was an emergent behavior in the DRL agents. The agents were not explicitly provided with a timer or biologically inspired neural structures to influence their behavior in order to be closer to biological systems. While previous studies have qualitatively reported emergent timing biases in deep neural networks (DNNs) \cite{deverett2019interval,roseboom2019activity}, to the best of our knowledge, this is the first study to suggest that DRL agents may exhibit temporal interference effects in a dual-task paradigm similar to those observed in human timing research.

\section{Analysis of neural dynamics}

\begin{figure}[h!]
  \centering
  \includegraphics[width=\textwidth]{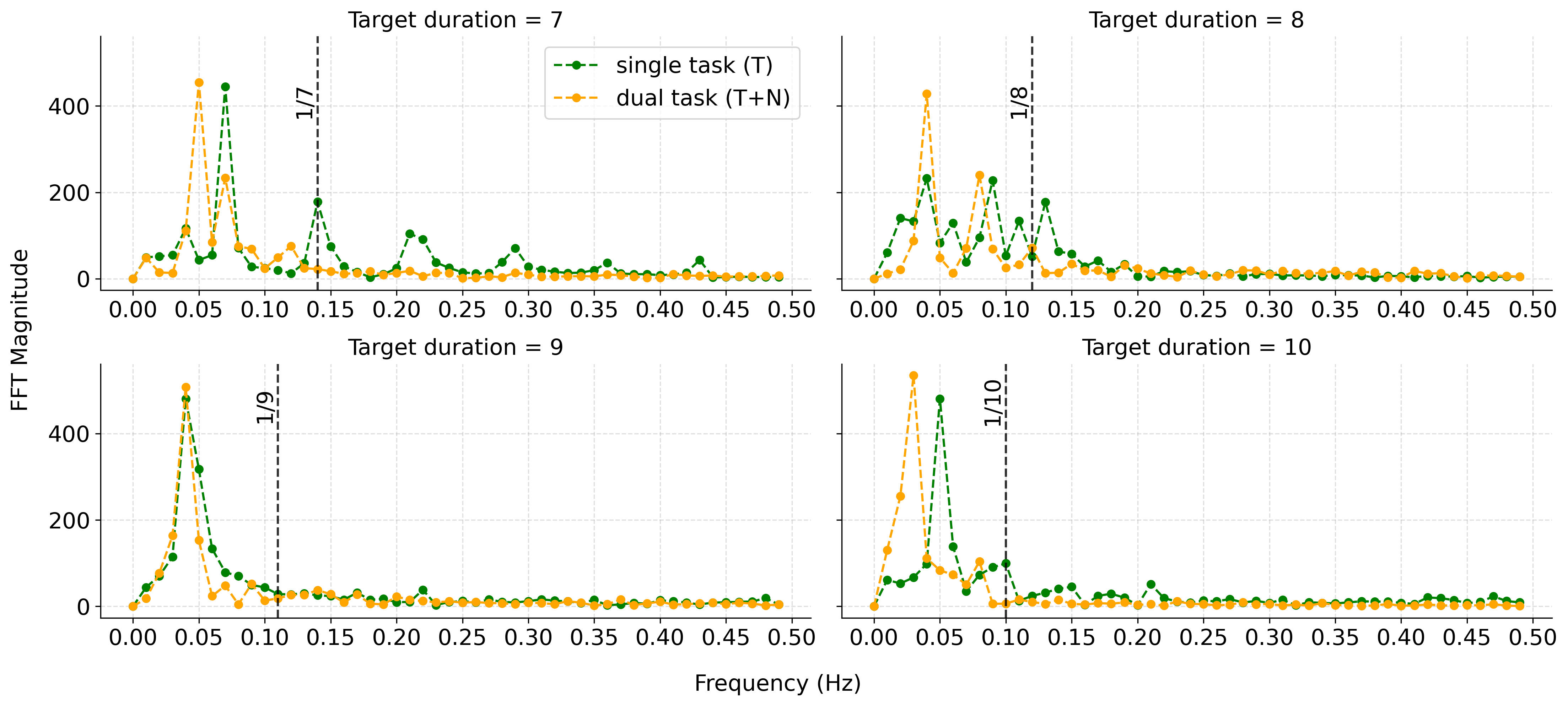}
  \caption{A fast fourier transform (FFT) of the first principal components of the LSTM hidden state activations (across first 100 time steps) for the single task (T) (in green) and the dual task (T+N) (in orange) agents across different durations. The black dotted line marks the target interval frequency. }
  \label{fig:fft_across_durations}
\end{figure}

% why lstm
An obvious place to start exploring the neural dynamics corresponding to temporal processing was the long short-term memory (LSTM) layer of the DRL agents. This is because, LSTMs, a type of recurrent neural network, are known to capture temporal dependencies in their input \cite{hochreiter1997long}. Additionally, in another variant of the DRL agents (not described in this study) where the LSTM layer was removed, the agent, while still able to successfully perform the single task (T), did not exhibit any time-keeping behavior. Instead, the agent simply placed the onion in the oven and repeatedly performed the `Interact' with oven action until it could take the soup out. Hence, it solely relied on state changes to perform the task. Moreover, studies involving continuous-time recurrent neural networks (CTRNNs) \cite{maniadakis2009explorations} and deep recurrent neural networks (RNNs) \cite{lin2023temporal} have demonstrated that recurrent layers can encode temporal information through biologically plausible mechanisms such as oscillations, ramping activity, and time cells.

% In contrast, the agents equipped with an LSTM layer, moved around the grid after placing the onion in the oven. 
% which are biologically plausible artificial recurrent network
% what is the analysis 
% add pca reference,
A principal component analysis (PCA) of the LSTM layer's hidden state activations across first 100 time steps revealed complex oscillatory dynamics with multiple peaks within each cycle (see Appendix). These oscillations were reset at the onset of each new trial (after soup delivery). Thus, it clearly carried information about the start and end of a trial. However, no evidence of a timer or any type of timing mechanism was observed in this analysis. Interestingly, previous work involving a DRL agent performing a dedicated time reproduction task showed counter-like neural activities in the PCA of the LSTM layer \cite{deverett2019interval}. The absence of such a dedicated timing mechanism may be attributed to the timing task being embedded in the soup delivery task and no immediate reward provided for correctly producing the target interval. 
% but previous work shows counter so it may be specifc to this task 

A fast fourier transform (FFT) was applied on the first principal components to disentangle the different oscillation patterns (see Figure \ref{fig:fft_across_durations}). For target durations 7 and 10, the single task (T) agent exhibited some oscillations with frequencies equal to or greater than the target duration, suggesting accurate timing or underproduction. In contrast, the dual task (T+N) counterparts showed oscillations with frequencies lower than the target duration, suggesting overproduction. However, no clear link can be established between these peaks and an intrinsic time-keeping mechanism. Thus, further analysis is needed to explain the neural dynamics underlying the observed behavioral differences.

\section{Discussion}
% % behavioural, neural and combine both 
% In this study we showed that a DRL agent exhibits interference in time production due to presence of a concurrent task. This is in line with findings in human studies. Anlaysis of the LSTM neural activations did not reveal any evidence of a dedicated counter or clock-like mechanism. Intrinsic timign is expected in DRL agent becuase we did not encode an explicit timer and considerign the complexity of such agents a number of tasks dynamics are integrated in a latent space. However, the behavioural simialrity between the agent and human studies supports the idea that timing is intrinsic in humans and neurons are shared between timing and other resources. This is just a speculation becuase two systems having simialr behaviour do not nececarry have the same underlyign mechansisn. 

% LSTM has some role to play in time-keeping. No explicit separation in dual task observed inside the lstm layer. The timing interuption by a secodnary task as been shown. However, the reason for this delay is still unclear. Taking inspiration from biological explanation it could be The analysis suggests that this could be due to integrated processing of both tasks. however, further exploration is needed to find a conclusisve reason for this. Maybe it would help aid our understand of cognitve temporal processing. 

% context, bigger picture and lr 
This study aims to contribute to recent research interest of drawing parallels between DNNs and biological systems to facilitate a better understanding of both \cite{barrett2019analyzing,kanwisher2023using,cichy2019deep}.While such research is extensively carried out in vision and audio, studies on temporal processing remain limited \cite{deverett2019interval, lin2023temporal}. Prior research has examined time-keeping mechanisms in DRL agents, revealing biologically plausible features such as ramping cells, time cells, or timing biases. The current study adds to this line of work by demonstrating new biologically plausible behavior in DRL agents.  

This study has several limitations, a few of them are as follows. First, the reward structure could be improved by using a delayed reward in the dual task setting. This would better align with corresponding human timing studies and help to further verify the biological similarities in the observed behavior. Second, the study was limited to only four target durations, including a wider range could further substantiate the findings. Third, the analysis of neural dynamics focused exclusively on the LSTM layer and was preliminary in nature. Future work could extend this by incorporating other layers from the DRL agent's architecture.  

% Finally, separately analysing the action and value networks may provide deeper insights into the strategies employed by the agents.

% This study has several limitations. Exploring different reward structures, like including a delayed reward in the dual task instead of an immediate one could help to furhter verify the behaviour obersevation. Immediate reward may not be comparable with human studies of dual task. The study is limited to only four durations, exploration with other durations can provide further conclusive evidence of behavioural simialirties in dual task. Exploring other parameter like varying gammas to further confirm the results. We have focussed on the lstm layer only.  The analysis of LSTM hidden state neurons is preliminary and has scope of further improvement either by including more durations or by considering the cell states in conjuction to the hidden states. Separating the action and value network could also provide useful insights on the underlying strategy used by the agents. 

\bibliographystyle{plain}  
\bibliography{references}

\section{Appendix}

\subsection{RL environment}
The agent state consisted of a spatial array with different integers indicating the different objects. The oven state whether on or off, agent state whether carrying onion or carrying soup were represented as one-hot encoding through different channels of the input array. Thus, while the agent received a change in state each time the oven was on, it did not receive any indication that the onion soup was ready (i.e., the target interval had passed). There was no upper limit on when the agent could take out the soup. Once taken out, the oven state changed to off and the agent input state updated.  The agent could take six actions in total. Five of which corresponded to navigating in the grid (`wait', `up', `down', `left' and `right') and one action to `interact' with the different objects by standing adjacent to it. 

% \begin{figure}[htbp]
%     \centering

%     \begin{minipage}[t]{0.48\textwidth}
%         \centering
%         \vspace{-7.5cm}
%         \includegraphics[width=\textwidth]{figures/overcooked_singleagentv2.drawio.pdf}
%         \vspace{1cm}
%         \caption{Grid environment for single task (T). Icons have been designed using resources from Flaticon.com}
%         \label{fig:grid_singletask}
%     \end{minipage}
%     \hfill
%     \begin{minipage}[t]{0.48\textwidth}
%         \centering
        
%         \includegraphics[width=\textwidth]{figures/overcooked_singleagentv1.drawio.pdf}
        
%         \caption{Grid environment for dual task (T+N). Icons have been designed using resources from Flaticon.com}
%         \label{fig:grid_dualtask}
%     \end{minipage}

% \end{figure}

\subsection{DRL agent architecture}

The agent architecture consisted of a convolutional neural network (CNN) layer designed to process spatial information from the 5x3 grid environment, primarily utilizing 1x1 convolutions. This was followed by a long short-term memory (LSTM) layer comprising 256 hidden units, which captured temporal dependencies. Subsequently, a multilayer perceptron (MLP) with 64 hidden units further processed the extracted features

\subsection{PCA of LSTM hidden states}

\begin{figure}[h!]
  \centering
  \includegraphics[width=\textwidth, height=5cm]{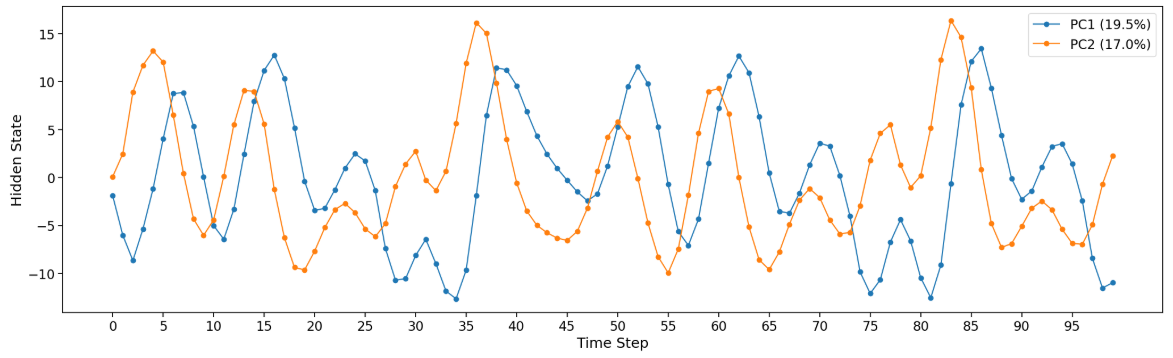}
  \caption{PCA of LSTM hidden state activations across 100 time steps for single task (T) agent }
  \label{fig:pca_singletask}
\end{figure}

\begin{figure}[htbp]
  \centering
  % \vspace{-8cm}
  \includegraphics[width=\textwidth, height=5cm]{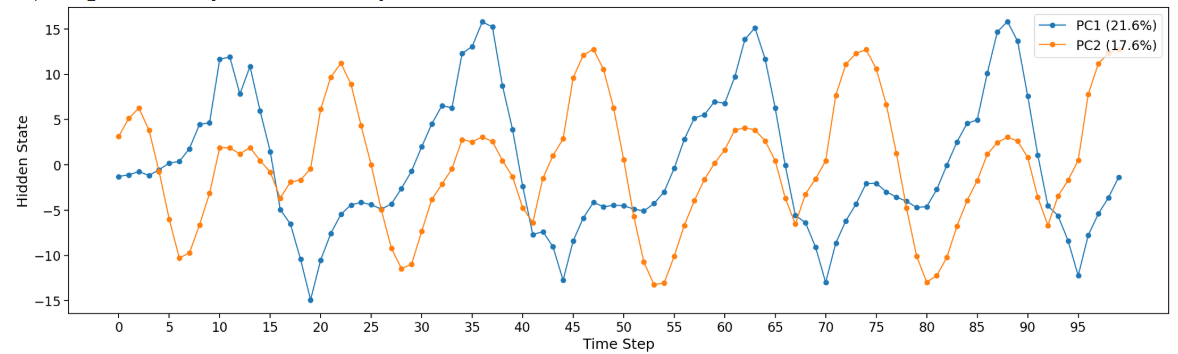}
  \caption{PCA of LSTM hidden state activations across 100 time steps for dual task (T+N) agent }
  \label{fig:pca_dualtask}
\end{figure}

\end{document}